\def\year{2019}\relax
\documentclass[letterpaper]{article} 
\usepackage{aaai19}  
\usepackage{times}  
\usepackage{helvet}  
\usepackage{courier}  
\usepackage{url}  
\usepackage{graphicx}  
\usepackage[utf8]{inputenc} 
\usepackage[T1]{fontenc}    
\usepackage{hyperref}       
\usepackage{url}            
\usepackage{booktabs}       
\usepackage{amsfonts}       
\usepackage{nicefrac}       
\usepackage{microtype}      

\usepackage{subfigure}
\usepackage{graphicx}
\usepackage{listings}
\usepackage{color}
\usepackage{xcolor}
\usepackage{nicefrac}
\usepackage{bibunits}
\usepackage{amsmath, amssymb}
\usepackage{float}
\usepackage{algorithm2e}

\usepackage{graphicx}
\usepackage{nicefrac}
\usepackage{array}

\usepackage{todonotes}

\newcommand{\DAgger}{\text{DAgger}}

\newcommand{\vizdoom}{\text{ViZDoom}}

\newcommand{\monte}{\text{Montezuma's Revenge}}

\newcommand{\MyWayHome}{\text{MyWayHome}}

\usepackage[font={it},labelfont={it}]{caption}

\newcommand{\M}{\text{M}}
\newcommand{\pts}{\text{pts}}
\newcommand{\adv}{\text{adv}}
\newcommand{\expert}{\text{expert}}
\newcommand{\AdwaC}{\text{A2C}}
\begin{document}
\title{Expert-augmented actor-critic for \vizdoom\ and \monte}
\author{
Michał Garmulewicz\\ 
University of Warsaw 
\And
Henryk Michalewski,\  Piotr Miłoś\\
University of Warsaw \\
deepsense.ai
}
\maketitle
\begin{abstract}
We propose an expert-augmented actor-critic algorithm, which we evaluate on two environments with sparse rewards: \monte\ and a demanding maze from the \vizdoom\ suite. 
In the case of \monte, an agent trained with our method achieves very good results consistently scoring above 27,000 points (in many experiments beating the first world). With an appropriate choice of hyperparameters, our algorithm 
surpasses the performance of the expert data. In a number of experiments, we have observed an unreported bug in \monte\ which allowed the agent to score more than $800,000$ points. 

\end{abstract}

\section{Introduction} \label{intro}

Deep reinforcement learning has shown impressive results in simulated environments \cite{dqn,a3c}. However, current approaches often fail when rewards are sparse as the cost of random exploration increases rapidly with the distance to rewards. This substantially restricts the scope of possible real-world applications, e.g. rewards in robotics are often assigned only once a task is completed and thus are binary and sparse \cite{her,lev-demo}. Additionally, random exploration potentially leads to safety issues. 

One way to improve the efficiency of exploration is to utilize expert data. The simplest technique, behavioral cloning, often suffers from compounding errors when drifting away from supervisor's demonstrations, see also Figure \ref{fig:pitfall} in Section \ref{discussion}. This can be mitigated by iterative methods like DAgger \cite{dagger} at the cost of a cumbersome data collection process. Recent work \cite{DQfD} has analyzed performance of behavioral cloning on Atari 2600 games. In the case of Montezuma's Revenge, cloning of near-perfect human demonstrations (a score above 30,000 points) led to a policy which scores merely 575 points.

We aim to address the shortcomings of cloning methods by combining the ACKTR algorithm \cite{acktr} with supervised learning from the expert data. Our method is easy to understand and implement. The core of our contribution can be expressed by a single formula, see Equation \eqref{eq:mainEquation}. The expert data is intended to guide the agent's exploration. 
We evaluate our method on two environments with sparse rewards: \monte\ and a demanding \MyWayHome\ maze from the \vizdoom\ suite 
\footnote{Please see videos of \monte\ gameplay (\url{https://bit.ly/2wG8Zh5}), exploiting an unreported bug in \monte ~(\url{https://bit.ly/2PC7cRi}) and \vizdoom\ MyWayHome gameplay (\url{https://bit.ly/2wMlxDL})}. Interestingly, our method is able to achieve stronger performance than the expert data, see Figure \ref{fig:montezumaresults}. 


We describe our method in Section \ref{expert-acktr}. In Section \ref{setup} we describe environments, training procedures, expert data and neural architectures. In Section \ref{experiments} we summarize our experiments. Section \ref{discussion} contains conclusions and suggestions of further work. 
Code for experiments presented in this paper is publicly available: \url{https://github.com/ghostFaceKillah/expert}.

        \begin{figure*}[h]
            \centering
                \includegraphics[scale=0.75]{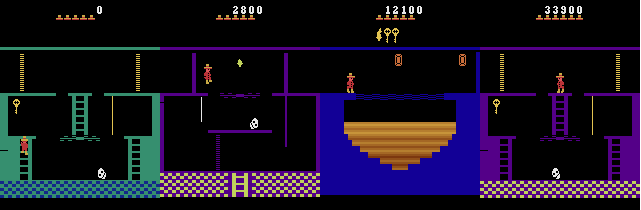}
                \caption{\monte\ - four screenshots from various stages of a gameplay. The room on the right marks the beginning of the second world. The agent reaches this room after passing through all the rooms in the first world. For the map of the first world see \url{https://atariage.com/2600/archives/strategy_MontezumasRevenge_Level1.html}, for the map of the second world see \url{https://atariage.com/2600/archives/strategy_MontezumasRevenge_Level2.html}. }
                \label{monte-screen}
        \end{figure*}

\section{Related work} \label{related-work}

\paragraph{Deep reinforcement learning}
The celebrated paper \cite{dqn} has sparked rapid development of deep reinforcement learning. The DQN algorithm presented there for the first time achieved (super)human performance in many Atari games. Soon after, other algorithms were developed e.g. A3C \cite{a3c} and TRPO \cite{trpo}. In this paper we employ the Actor-Critic using Kronecker-Factored Trust Region (ACKTR) algorithm \cite{acktr}. ACKTR is a policy-gradient method that utilizes natural gradient techniques \cite{natural-gradient} which have proved to be successful in increasing speed and stability of learning. 

\paragraph{Expert data in reinforcement learning.} A vast body of literature has been devoted to imitation learning. 
As the learned policy influences the future states upon which it will be evaluated, errors compound and make this method impractical for tasks with long time horizon \cite{ross-pre-dagger} (e.g. behavioral cloning in Montezuma's Revenge \cite{DQfD}  yielded lackluster performance). The seminal work \cite{dagger} describes the \DAgger\ algorithm that alleviates this issue, however at the cost of the iterative expert data collection process. In our method, we collect a fixed set of expert data, which is reused in the whole experiment.

\begin{table*}[t]
\begin{tabular}{
>{\raggedright\arraybackslash}p{4.7cm} 
>{\raggedleft\arraybackslash}p{0.8cm} 
>{\raggedleft\arraybackslash}p{0.8cm} 
>{\raggedleft\arraybackslash}p{0.8cm} 
>{\raggedright\arraybackslash}p{8cm} 
}
\toprule       
\textbf{Approach} & \textbf{Mean score} & \textbf{Max score} & \textbf{Trans. $\times 10^6$ } & \textbf{Methods used} \\
\midrule
ExpAugAC {\tiny (our method)} & 27,052   & 804,900 &  200 & {\tiny Expert loss term based on batches of prerecorded data} \\
DQfD \cite{DQfD}            & 4,740 & -  &  200     & {\tiny 750k batches of expert pretraining, 3 additional loss terms, prioritizing expert data replay} \\
Behavioral cloning \cite{DQfD}      & 575 & -  & 24 & {\tiny \text{--}} \\
Ape-X DQfD \cite{apex-dqfd}      & 29,384 & - &  2,500 & {\tiny Methods from DQfD, temporal consistency loss, transformed Bellman operator}  \\
Playing hard YT \cite{youtube}& 41,098 & - & 1,000  & {\tiny Auxiliary reward encouraging imitation of videos of expert gameplays } \\
Learning MR from a Single Demonstration \cite{openai-monte} & -  &  74,500 & - & {\tiny Decomposing task into a curriculum of shorter subtasks, assumes ability to set env. state }  \\
Unifying Count-Based Exploration \cite{unifying-count-based} & 3,439 & 6,600 & 100  &   {\tiny Exploration-based auxiliary reward, mixing Double Q-learning target with MC return}   \\
\bottomrule      
\end{tabular}
\caption{The state of the art for \monte.}
\label{table:sota}
\end{table*}

Value-based deep reinforcement learning methods have been combined with learning from demonstrations \cite{DQfD,apex-dqfd}. In these works, the learning algorithm is altered to ensure that the expert data added to the replay buffer is used efficiently. For example, no action is allowed to have a higher Q-value than expert's action, the $n$-step return loss is used to propagate expert information and expert demonstrations are sampled more often. A different approach, combining DQN with expert data classification loss has been presented in \cite{FewExpert}. However, authors do not test their method on \monte\ or \vizdoom. In our work, we only use one expert imitation loss. Formulation of this loss is motivated by the policy gradient theorem, see Formula \eqref{eq:mainEquation}. 

Methods combining policy gradient algorithm with expert demonstrations have been successfully applied to problems such as complex robotic manipulation \cite{complex-dexterous-manipulation} or dialogue management \cite{dialogue}. In the case of Atari environments, in \cite{Zhang} expert data is used to pre-train policies that are later passed to a policy gradient algorithm. A substantial effort has been put into integrating replay buffers into actor-critic algorithms including \cite{wawrz2009,wawrz2013,acer}.

\paragraph{\monte}
Significant research attention has been dedicated to \monte\, as it is regarded 

as a testing ground that challenges exploration abilities of a given learning algorithm. Table \ref{table:sota} summarizes the current state of the art, including methods using expert data. Efforts have been made to tackle this environment by adding natural language instructions \cite{monte-lang} (avg. 3,500 pts using 60\M\ transitions), extending model with intrinsic curiosity rewards in \cite{monte-count-based} (avg. 3,700 \pts\ using 150\M\ transitions) or introducing hierarchy into the model \cite{feudal} (avg. 2600 pts using 1000M frames). Numbers in the parenthesis represent scores and amount of training samples reported in these papers. Combining hierarchical learning and imitation learning has been demonstrated to be effective \cite{hier}, although the work focuses only on the first screen of \monte. 

\paragraph{Doom}

\vizdoom\ \cite{vizdoom} is a popular suite of reinforcement learning environments based on the first-person shooter game Doom. Impressive results have been obtained for various tasks in these environments. These tasks include in particular navigation \cite{semi-topo,lstm-iter-nets,neural-map} and multi-player combat \cite{learning-to-act,arnold,actor-critic-curriculum,deep-successor}. 
In \cite{td-or-not-td,regret-minimization,rl-squared,transfer-drl-3d} the \vizdoom\ suite was used as a benchmark in other reinforcement learning experiments. 

Curiosity-based intrinsic rewards have been used in \cite{curiosity} to increase efficiency of exploration in \vizdoom. Authors of  \cite{curiosity} report that in the navigation task \MyWayHome\ (see Section \ref{setup} for a description) the standard actor-critic approach completely fails, while their method solves the task in 70\% of the cases after using 10M transitions for training. Our algorithm always solves the task after training on 5M transitions. To the best of our knowledge, our method is the first one to use expert data in \vizdoom.

\section{Expert-augmented ACKTR} \label{expert-acktr}

        \begin{figure*}[ht]
            \centering
                \includegraphics[scale=0.7]{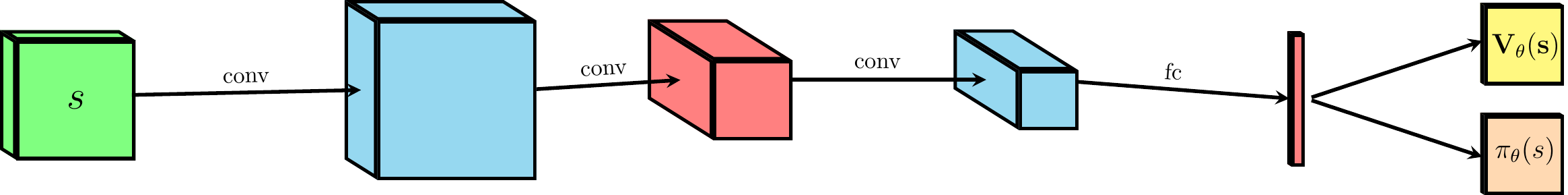}
                \caption{Network architecture: stacked input frames are passed through a convolution layer with 32 filters size of 8 and stride 4, followed by 64 convolution filters of size 4 and stride 2, followed by 64 convolution filters of size 3 and stride 1. The output is fed into a fully connected layer with 512 units. Next, two heads are attached. The first one with a single neuron for the value function approximation and the second one with the number of neurons matching the dimension of the action space. After applying the softmax function the second head outputs the policy. In all layers, except the top heads, we use the ReLU non-linearity.}
                \label{fig:architecture}
        \end{figure*}
        
Reinforcement learning is formalized in the framework of Markov decision processes (MDP). In this setting at each timestep $t$, the agent observes a state $s_t \in S$ and chooses an action $a_t \in A$ according to policy $\pi$, which is a (potentially stochastic) mapping from a given state to actions. 

The environment then returns reward $r(s_t, a_t)$ and evolves to the next state $s_{t+1}$ according to a transition probability $P(s_{t+1}|s_t, a_t)$.
The goal of reinforcement learning is to find a policy within some class which maximizes 
\[\mathcal{J}(\pi) = \mathbb{E}_{\pi}[R_t],\] where \[R_t = \sum_{i=0}^{\infty} \gamma^i r(s_{t+i}, a_{t+i}),\ \gamma\in (0,1),\] is the future discounted reward. In our applications the set of available neural network policies is parametrized by $\theta$. Policy-based methods \cite{generalized-advantage,sutton-pg,williams-pg} are inspired by the Policy Gradient Theorem which states that
 \begin{equation*}
 \nabla_\theta \mathcal{J}(\pi_\theta) =
 \mathbb{E}_{\pi_{\theta}} \left[
         \sum_t \adv_t
          \nabla_\theta \left( \log \pi_\theta (a_t | s_t)\right)
\right], 
 \end{equation*}
where \[\adv_t = R_t - b(s_t)\] and $b$ is an arbitrary function of the state, called a {\em baseline} function. In the context of A2C and ACKTR algorithms we take the approximate state-value function $V_\theta$ as a baseline. Thus we aim to optimize the following loss \[L^{\text{A2C}} (\pi_\theta) = \sum_t L^{\text{A2C}}_t(\pi_\theta),\] where \[L^{\text{A2C}}_t(\pi_\theta) =  \mathbb{E}_{\pi_{\theta}}
        \left[-\text{adv}_t
         \log \pi_\theta (a_t | s_t)
        + \frac{1}{2} (R_t - V_\theta(s_t))^2\right].\]

The main algorithmic contribution of our paper consists in adding an extra term to the loss function to accommodate the expert data: 

\begin{equation}\label{eq:mainEquation}
 L(\pi_\theta) =
 L^{\AdwaC}(\pi_\theta) - \frac{\lambda_\expert}{k}
    \sum_{i=1}^k
       \text{adv}^{\text{expert}}_i
        \log \pi_\theta (a^{\text{expert}}_i | s^{\text{expert}}_i),
\end{equation}
where $a^{\text{expert}}_i$ and $s^{\text{expert}}_i$ are sampled from a fixed dataset of expert rollouts.

We consider three variants of the expert advantage:
\[ 
\begin{split}\textit{reward:\ } \text{adv}^{\text{expert}}_t = & \sum_{s\geq 0}  \gamma^s r_{t+s}^{\text{expert}} ,\\
 \textit{critic:\ }   \text{adv}^{\text{expert}}_t = &
        \left[ \sum_{s\geq 0}  \gamma^s r_{t+s}^{\text{expert}}- V_\theta(s_t) \right]_{+} ,\\  \textit{simple:\ }  \text{adv}^{\text{expert}}_t = & \ 1.  \end{split}\]
where $[x]_+ = \max(x, 0)$. We follow the ACKTR algorithm which estimates the natural gradient direction associated with the current policy, see Algorithm \ref{listing} for additional details.

\section{Environments, training and architectures} \label{setup}

\begin{figure*}
\subfigure[]{\includegraphics[scale=0.7]{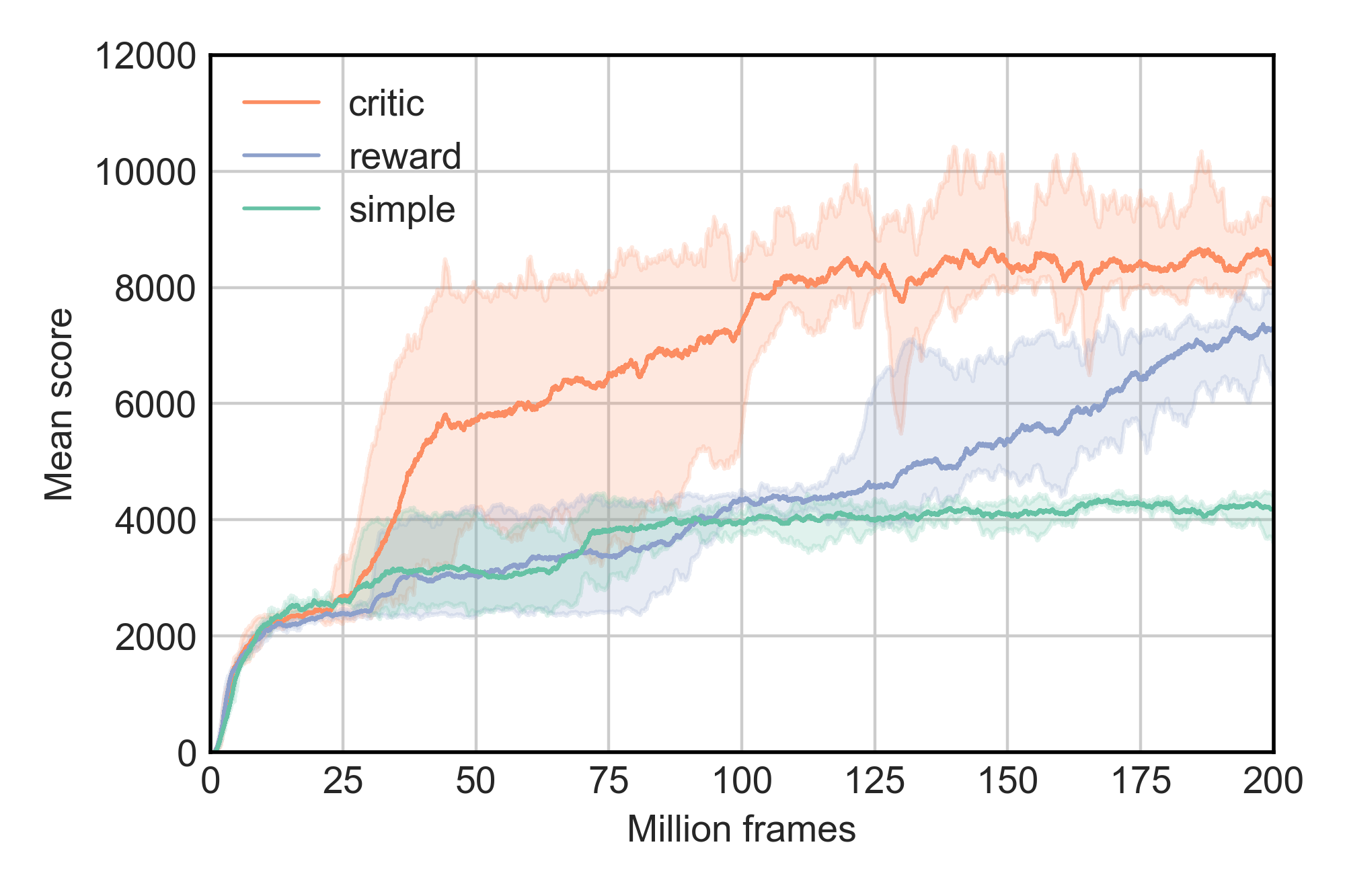}}
\subfigure[]{\includegraphics[scale=0.7]{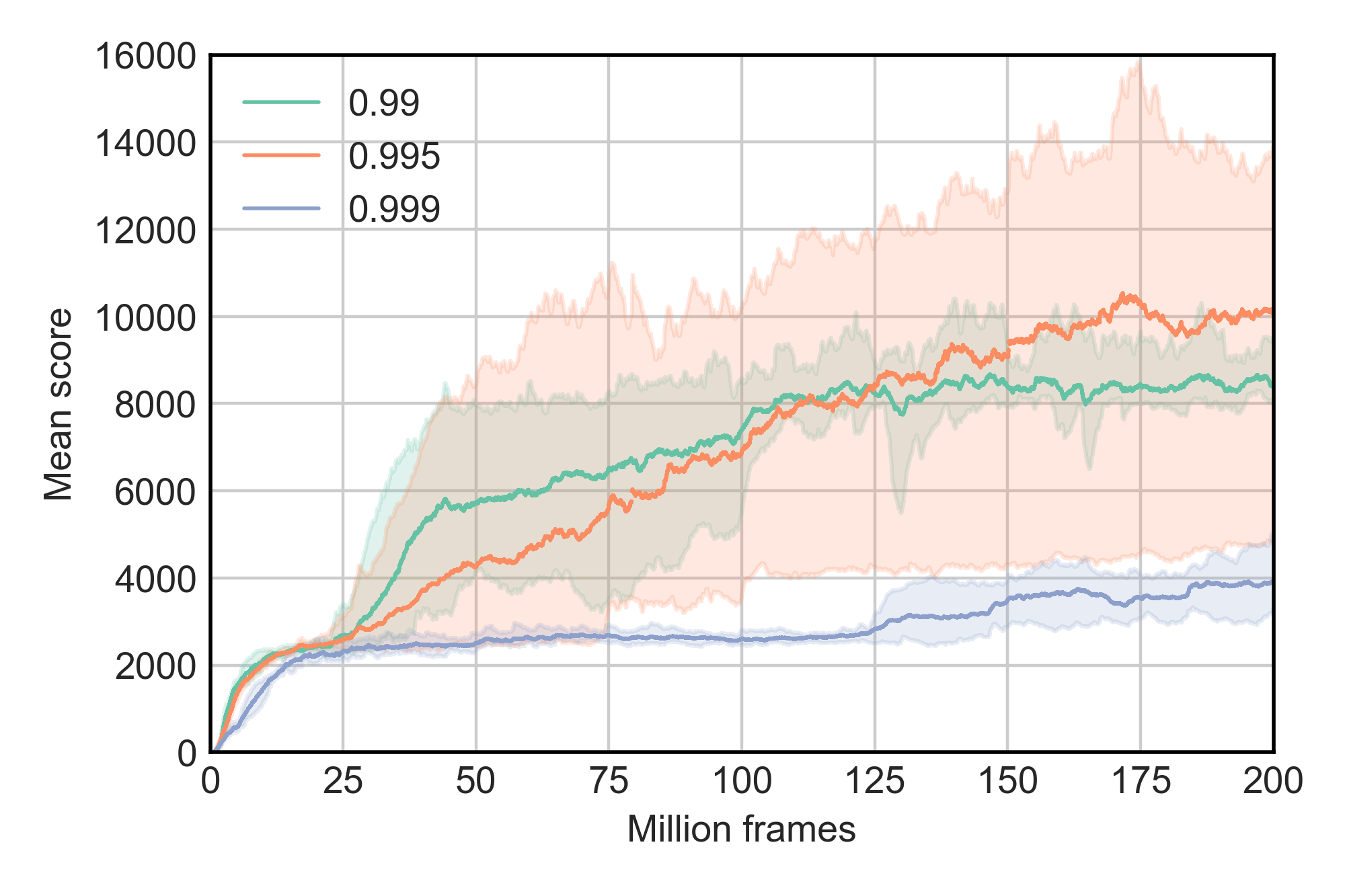}}
\subfigure[]{\includegraphics[scale=0.7]{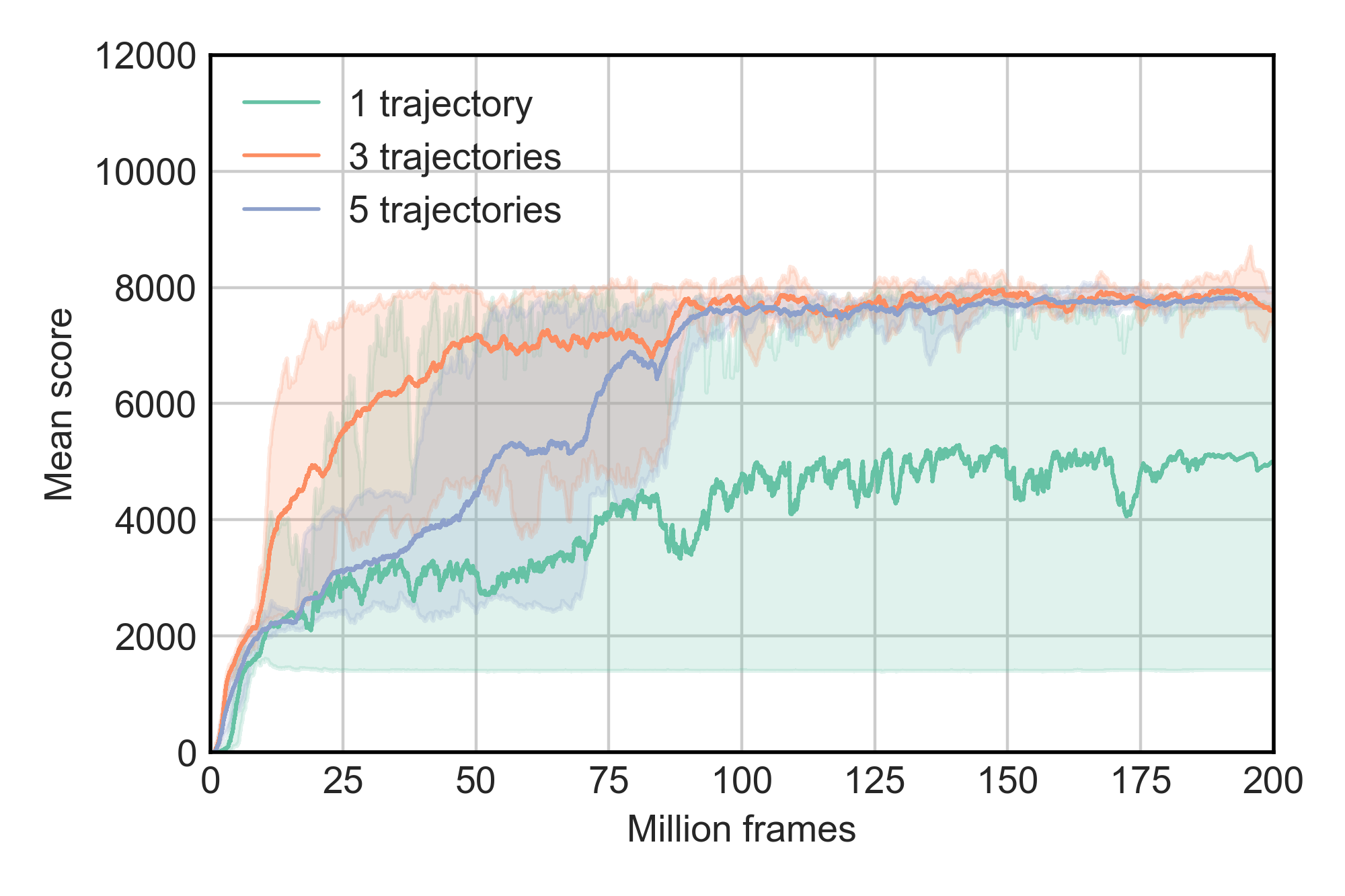}}
\subfigure[]{\includegraphics[scale=0.69]{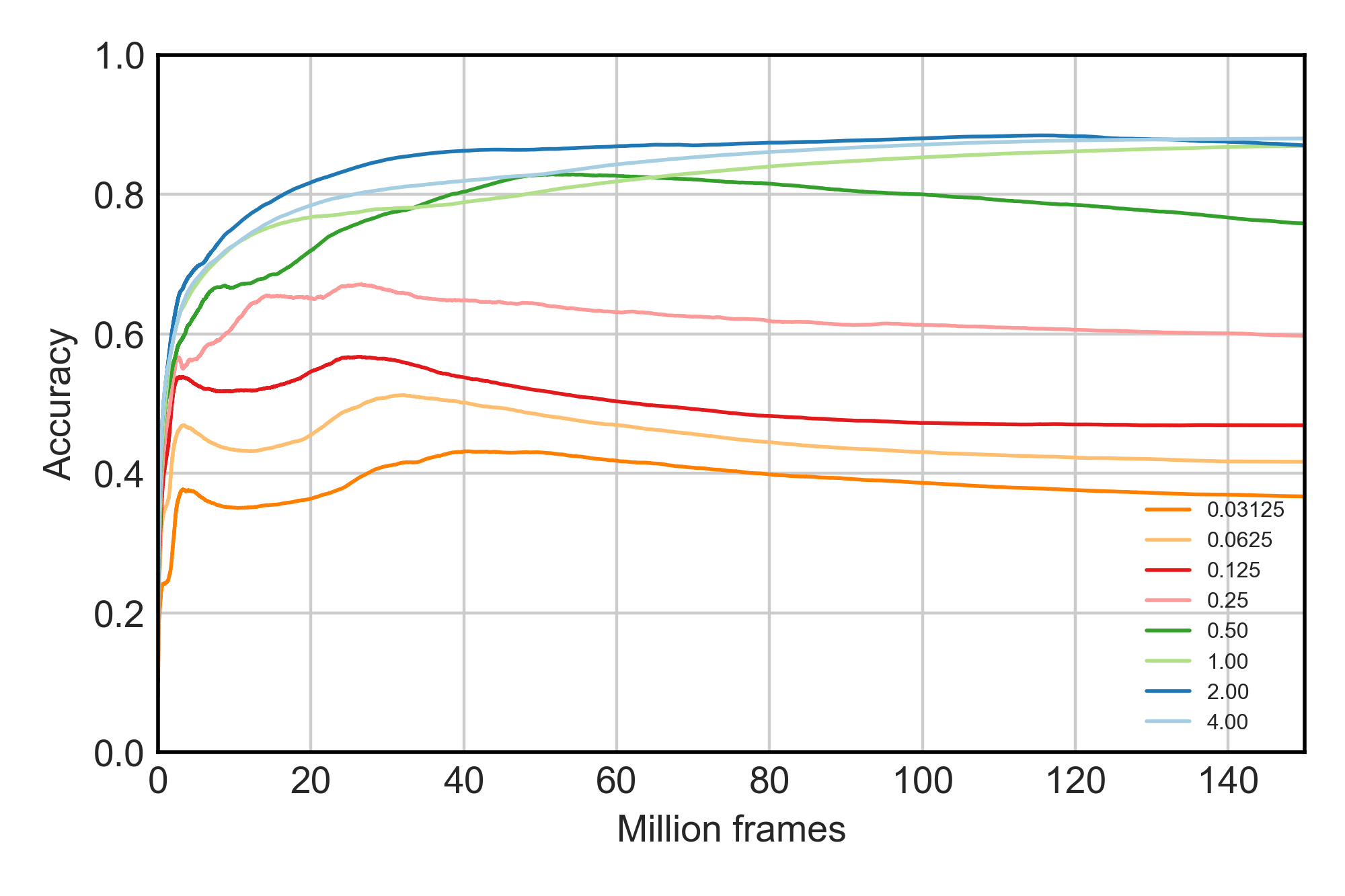}}
\subfigure[]{\includegraphics[scale=0.7]{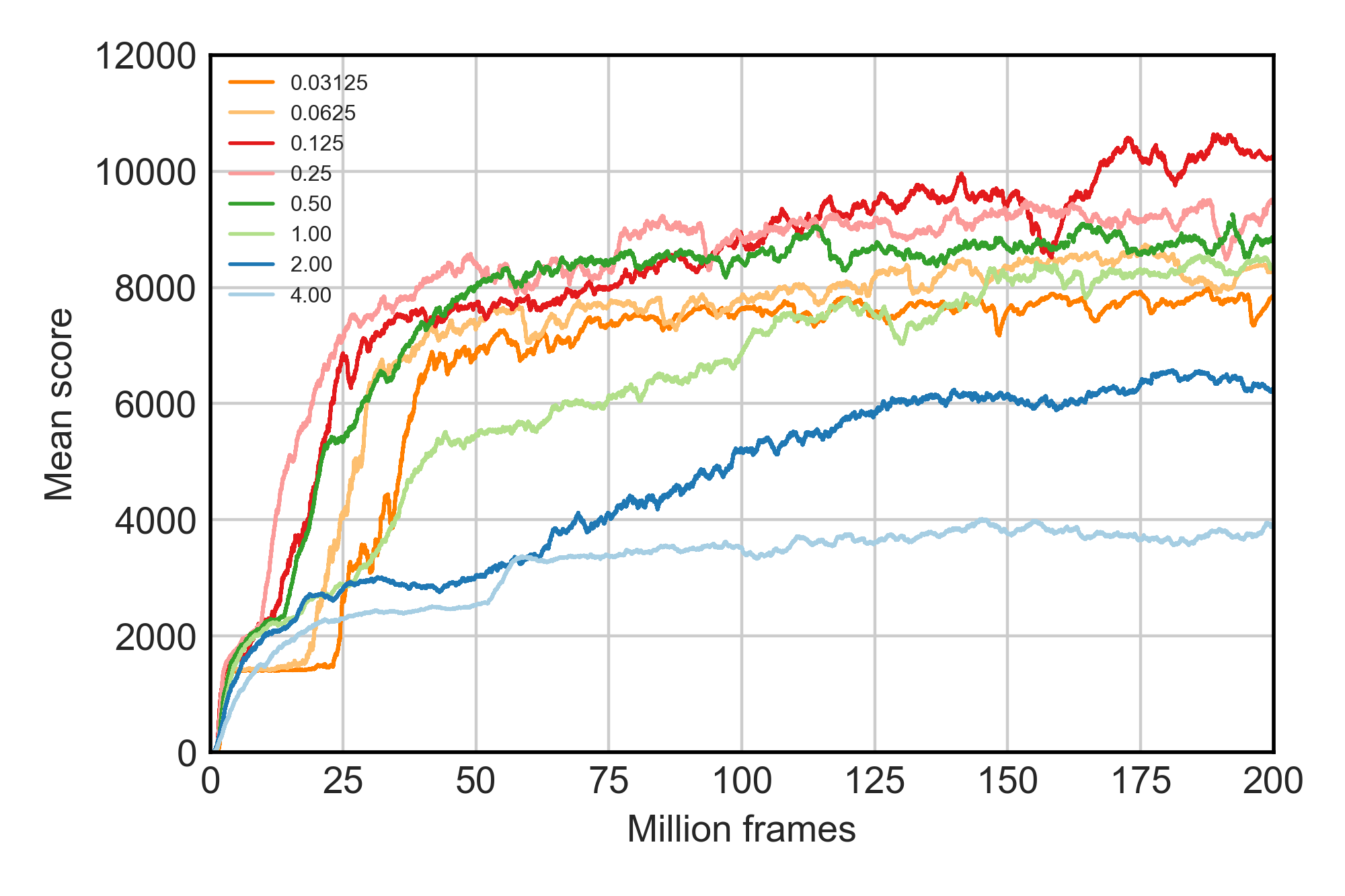}}
\subfigure[]{\includegraphics[scale=0.7]{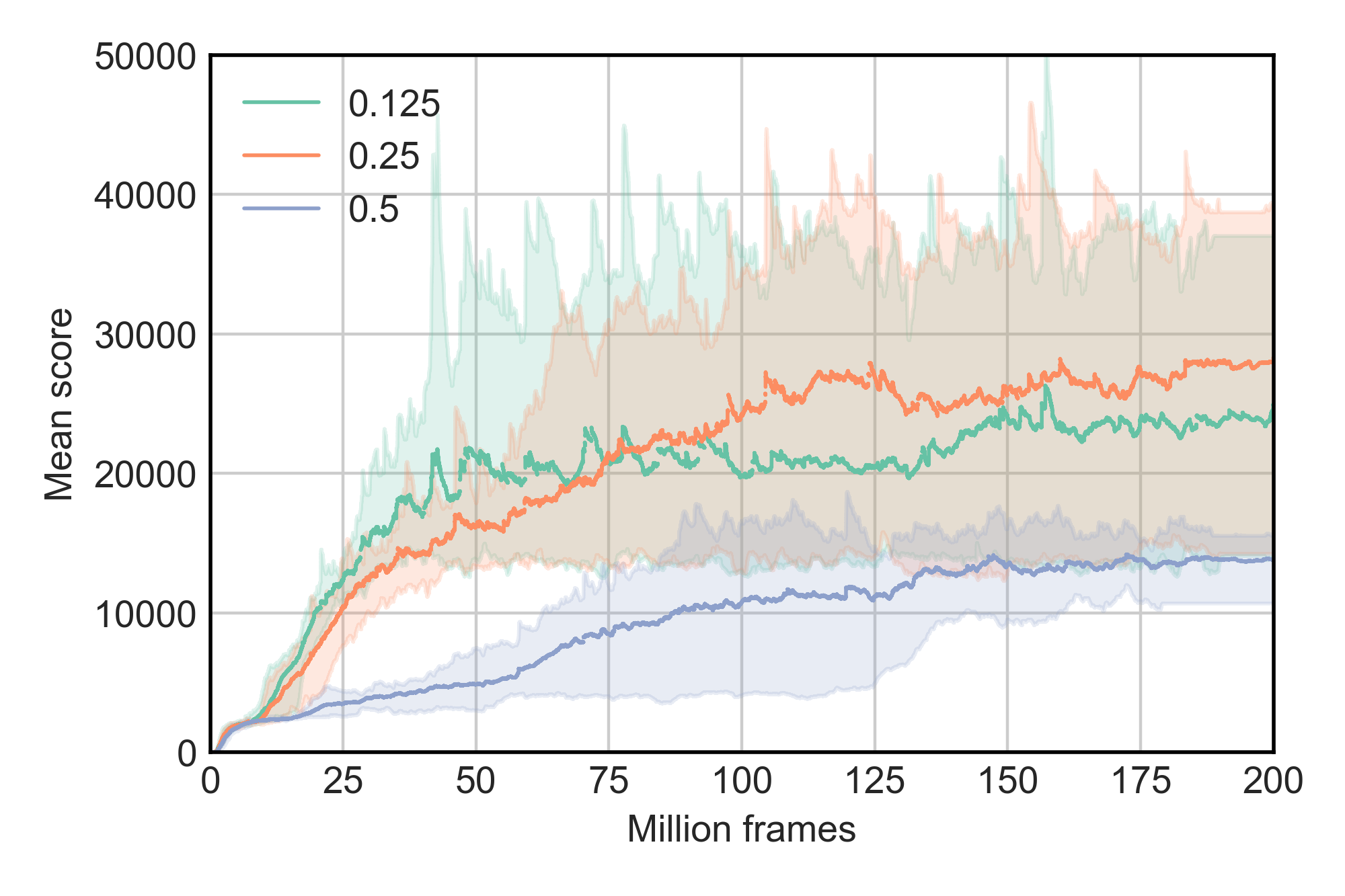}}
  \caption{\monte\ experimental results. See the top left plot for a comparison between various advantage estimators defined in Section~\ref{expert-acktr}.
  In experiments depicted in the top right plot we vary the discount rate $\gamma$. The sweet spot is around $\gamma=0.995$. We conjecture that even higher values could be beneficial, but would require better handling of the variance of empirical rewards. 
Plot in the center-left shows that smaller amount of expert data yields higher variance of results.   
In the bottom-left plot we test various choices of $\lambda_{\text{expert}}$ to assess utilization of the expert data. The performance peaks for the intermediate values. For small values expert's learning signal is too weak, while high values impair the final performance because the agent overfits to the expert data, as seen in the train set accuracy plot (center-right). In the experiments presented in the bottom-right plot we provide scores for optimized parameters, that is we apply $\gamma=0.995$, the critic advantage estimator and selected values of $\lambda_{\text{expert}}$.}
  \label{fig:montezumaresults}
  \end{figure*}

        \begin{figure*}[ht]
            \centering
                \includegraphics[scale=0.37]{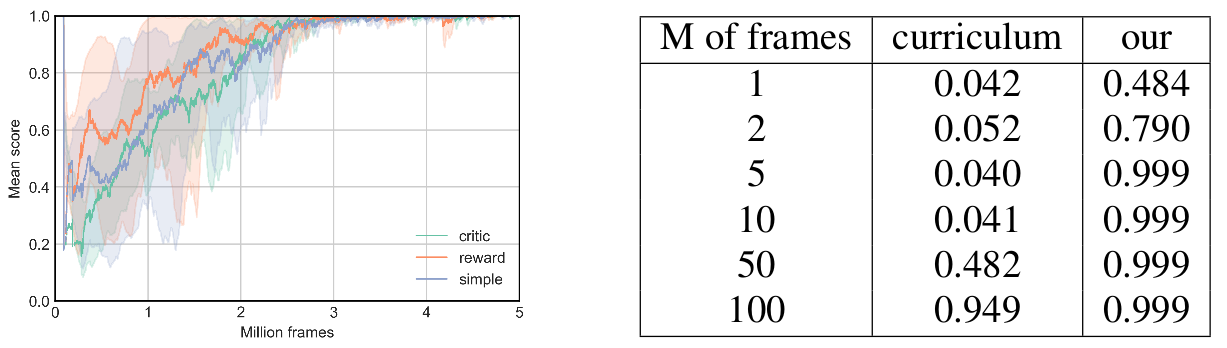}
  \caption{Left: In \vizdoom\ behavior is similar for all expert advantage estimators. Right: performance of a curriculum learning in \MyWayHome\ compared to our expert-augmented algorithm. The curriculum consists in re-spawning the agent in random locations. We experimentally verified that  the  ACKTR algorithm without the curriculum was unable to solve the \MyWayHome\ task. This echoes observations made for another actor-critic model-free algorithm in (PAED17). Our behavioral cloning experiments also failed to solve this task. The curiosity-based method described in (PAED17) achieves an average score $0.7$ after 10M of frames. }
  \label{fig:viz}
        \end{figure*}

\paragraph{Doom}

\begin{figure}[H]
\includegraphics[scale=0.35]{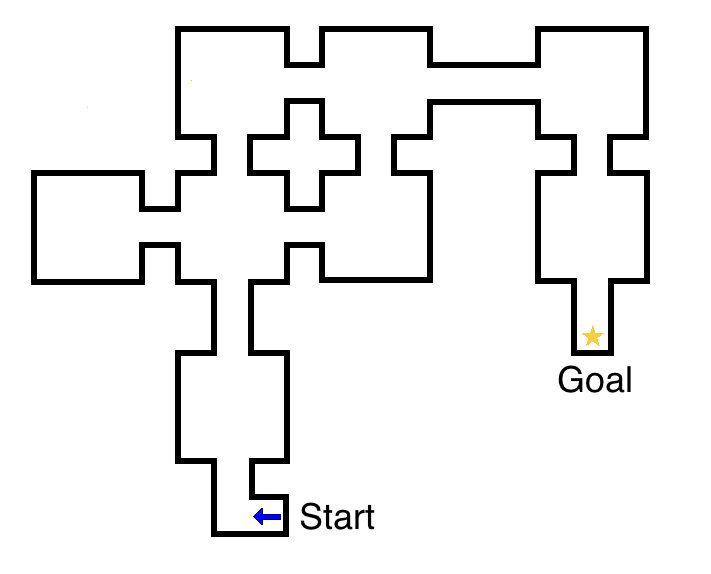}    
	\caption{The map of the \MyWayHome\ maze. The starting position and the goal are always located in the same place. See Figure \ref{doom-scene} for a screenshot from the environment.}
        \label{doom-map}
\end{figure}

\MyWayHome\ is a navigation task defined in the \vizdoom\ suite \cite{vizdoom}. In \MyWayHome\ the maze consists of 9 rooms, see Figure \ref{doom-map}. 

The agent receives reward $1$ if it reaches the {\em goal} location in less than 2100 time steps. There are not other rewards in the game. The {\em goal} is always located in the same place and 
the agent always starts in a fixed location far from the {\em goal}. 
The action space consists of three actions: {\em forward}, {\em turn left}, {\em turn right}. 

\begin{figure}[H]
\includegraphics[scale=0.36]{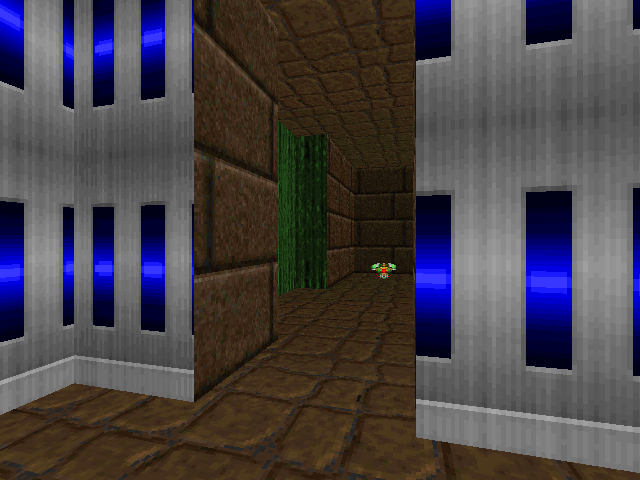}   
        \caption{An  observation from the \MyWayHome\ environment. The goal location is visible in the background in the next room. }
        \label{doom-scene}
\end{figure}

\paragraph{Montezuma's Revenge}
Montezuma's Revenge belongs  to the suite of Atari 2600 games \cite{ale}. The task in the game is to navigate in a labyrinth (see Figure \ref{monte-screen}), collect valuable items and avoid deadly obstacles.  \monte\ is known for its sparse rewards. For example, obtaining the only reward available in the first room (see the leftmost image in Figure \ref{monte-screen}) requires five ``macro actions'' such as ``jump across a gap'' or ``go down a ladder'', each spanning over dozens of environment steps.

\paragraph{Training details}
Training details are shared across \monte\ and \vizdoom's \MyWayHome\ environments. The input to the policy consists of the stacked last four frames which are downscaled to $84\times 84$ pixels and cast to grayscale.  We use Algorithm 1 with the learning rate $0.125$, 32 actors and the time horizon $T=20$. 

For the expert data we use $k=256$ (see Equation \ref{eq:mainEquation}). Following \cite{a3c}, given entropy $H$ we introduce an additional entropy regularization term: $\beta_{\text{entropy}}$ defined as 
$\nabla_{\theta} H(\pi_{\theta}(\cdot|s_t))$. 
We set 
$\beta_{\text{entropy}} = 0.001$.

Our implementation of Algorithm 1 is based on the implementation of the ACKTR algorithm provided in   \cite{baselines}. 





\begin{algorithm}
{\Large 
 \KwData{Dataset of expert transitions $ (s^{\text{expert}}_t, a^{\text{expert}}_t, s^{\text{expert}}_{t+1}, r^{\text{expert}}_t) $ }
 Initialize neural net policy $\pi_\theta$
 
 \For{$iteration \leftarrow 1$ \KwTo max steps}{
   \For{$t \leftarrow 1$ \KwTo T}{
    Perform action $a_t$ according to $\pi_\theta(a|s_t)$ \\
    Receive reward $r_t$ and new state $s_{t+1}$
   }

   \For{$t \leftarrow 1$ \KwTo $T$}{
    Compute discounted future reward estimator: $\hat{R}_t = r_t + \gamma r_{t+1} + \ldots + \gamma^{T-t+1}r_{T-1} + \gamma^{T-t}V_{\theta}(s_T)$ \\
    Compute advantage estimator: $\widehat{\text{adv}}_t = \hat{R}_t - V_{\theta}(s_t)$
   }
   Compute $\nabla$ of the A2C loss 
   $g_{A2C} =  \nabla_\theta 
   \frac{1}{T}
   \sum_{t = 1}^T  \left[
     -\widehat{\text{adv}}_t \log \pi_\theta(a_t | s_t)  +
     \frac{1}{2} (\hat{R}_t - V_{\theta}(s_t))^2
   \right]  $ \\
   Sample mini-batch of $k$ expert state-action pairs \\
   Compute expert advantage estimator $\text{adv}^{\text{expert}}_t$ for each state-action pair \\
   Compute $\nabla$ of the expert loss 
   $g_{\text{expert}} = \nabla_\theta \sum_{i=1}^{k} \left[
    -\text{adv}^{\text{expert}}_i \log \pi_\theta(a^{\text{expert}}_i | s^{\text{expert}}_i) \right]
   $ \\
   Update ACKTR inverse Fisher estimate \\
   Use ACKTR Kronecker optimizer with gradient $g = g_{A2C} + \lambda_{\text{expert}}g_{\text{expert}}$.
 }}
 \caption{Expert-augmented ACTKR. Pseudocode for one actor.}
 \label{listing}
\end{algorithm}

\section{Experiments} \label{experiments}

In this Section we empirically verify a number of properties of our algorithm. In particular (see Figure \ref{fig:montezumaresults} for a summary):
\begin{enumerate}
\item we check which of proposed advantage estimators offers the best performance, 
\item we check how the discount factor $\gamma$ influences the reward and what are consequences of decreasing the number of expert  trajectories,
\item we verify what is the potential for generalization: in the case of \monte\ we provided expert trajectories only from the first world of \monte, but our agents are capable of continuing the gameplay in the second world. For the map of the first world see \url{https://atariage.com/2600/archives/strategy_MontezumasRevenge_Level1.html}, for the map of the second world see \url{https://atariage.com/2600/archives/strategy_MontezumasRevenge_Level2.html}
\item we also grid the  $\lambda_{\text{expert}}$ coefficient with respect to the total reward and verify how $\lambda_{\text{expert}}$ is related to the accuracy on expert data.
\end{enumerate}


\subsection{\monte}
For a summary of the state of the art compared with our results see the Table on page 2,
for evaluation videos  see \url{https://bit.ly/2wG8Zh5}. 
Interestingly, our algorithm discovered a bug, see video \url{https://bit.ly/2PC7cRi}, which manifests through scores exceeding $800,000$ in some evaluation rollouts (these rollouts are excluded when we calculate the mean score). 
In all experiments reported in this Section we performed at least $5$ experiments with different seeds. Shades represent min-max ranges. In all experiments except for  Figure \ref{fig:montezumaresults} top-right and bottom-right, we set  $\gamma = 0.99$; in the top-left, top-right and center-left experiments we set $\lambda_{\text{expert}}=1$; in all experiments except for Figure \ref{fig:montezumaresults} top-left we use the critic advantage estimator and in all  experiments except for Figure  \ref{fig:montezumaresults} center-left we use all  available expert trajectories (14 trajectories). In the experiment presented in the bottom-right we selected the best performing parameters from previous experiments, that is the critic advantage estimator and $\gamma=0.995$. We also set the advantage estimator for expert-data  $\lambda_{\text{expert}}$ equal to $0.125, 0.25$ or $0.5$.

In order to address the question of generalization from the first \monte\ world to the second \monte\ world (see the third point at the beginning of this Section), we considered 6 runs for the best parameters ($\lambda_{\text{expert}}=0.125$, $\gamma=0.995$). Please note that all provided expert trajectories were limited to the first world. Let us choose a training run which achieves the median score at the end of training. The stochastic policy associated with this run receives average reward $27,052$ in $100$ evaluations. It completes the first world in $60\%$ of the cases and outperforms the best expert trajectory in $16\%$ of the cases.
In contrast, if we choose the policy not from the median, but from the best run, it resolves the first world in $88\%$ of the cases and outperforms the best expert trajectory in $86\%$ of the cases. Interestingly, in some rollouts (not taken into the average) our policies exploit an unreported bug in \monte\, occasionally scoring over $100,000$ points with a maximum of $804,900$ points.

\subsection{\vizdoom\ \MyWayHome}
In Figure \ref{fig:viz} we summarize an experiment which compares the efficiency of the model-free ACKTR algorithm with our expert-augmented method. The model-free approach without the expert data
does not work. However, the agent learns how to navigate the maze after a relatively mild simplification of the environment consisting in allowing the agent to re-spawn in random locations in the labyrinth, which can be seen as a curriculum learning. The agent manages to learn a consistent behavior in 100M actions. However, if we allow expert data, the agent reaches even a better performance already after 5M actions without any curriculum, see Figure \ref{fig:viz}.  


\section{Conclusions and future work} \label{discussion}
Based on experimental results we claim
that the algorithm presented in this work is a practical method of getting good performance
in cases when multiple interactions with the environment are possible and good
quality expert data is available. It could be particularly useful in settings such as  \monte, where
neither supervised learning from expert data nor random exploration yield good
results.

Using model-free reinforcement learning algorithms with expert data brings a number of issues which are worth addressing. In particular, our expert-augmented algorithm would benefit from an adaptive $\lambda_{\text{expert}}$ coefficient and accuracy rate targeting on expert trajectories  (from our experiments follows that for overall performance of the algorithm it would be most beneficial to keep accuracy between $0.6$ and $0.9$, see Figure \ref{fig:montezumaresults}). In our view, another interesting research direction would be to design a proxy metric which would identify situations requiring additional expert data. That is situations where we are far away from the expert trajectory and simultaneously the model-free algorithm does not have a capacity to find any rewards on its own, see Figure \ref{fig:pitfall}. One can regard this as an automation of the DAgger algorithm \cite{dagger}. 
\begin{minipage}{0.99\linewidth}
    \centering
    \begin{figure}[H]
        \begin{minipage}{0.45\linewidth}\includegraphics[scale=0.65]{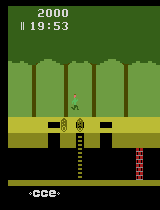}    
        \end{minipage}
        \hspace{0.08\linewidth}
        \begin{minipage}{0.45\linewidth}\includegraphics[scale=0.65]{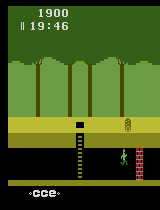}   
        \end{minipage}   
        \caption{Left: our agent in Pitfall! trained on an expert trajectory which achieves the perfect score (see a run \url{https://youtu.be/MX_e2UZLk1E} of our agent). The expert trajectory and our agent run only above the ground. Right: the agent falls into the lower level and is unable to return to the expert trajectory, see video \url{https://youtu.be/07327Eh6zIM}.}
        \label{fig:pitfall}
    \end{figure}
\end{minipage}

\medskip
We also leave  as future work the following other extensions of the experiments presented in this article:
\begin{enumerate}
\item Measure performance of the algorithm when the quality of expert data varies. 
\item Rewards in robotics are often assigned only once a task is completed and thus are binary and sparse \cite{her,lev-demo}. In our view it would be interesting to test methods presented in this paper on robotic object manipulation tasks or in a simulated car environment. 
\item Authors of \cite[Table 3]{apex-dqfd} report Bank Heist, Gravitar, Ms. Pacman, Pitfall! and Solaris as games where human experts are still stronger than AI even when expert trajectories are available in the process of training. In an on-going experiment, we assess the performance of our method in these environments. 
\item Using human knowledge in tasks routine for humans but difficult for artificial intelligence, such as driving, seems to be a particularly important application of methods presented in this paper. We test these ideas in an on-going experiment with the Carla simulator \cite{carla}.
\item In our experiments, the discount factor $\gamma=0.995$ performed optimally, see Figure \ref{fig:montezumaresults}. Can we get further to higher values of $\gamma$?
\end{enumerate}


  
\bibliography{short}
\bibliographystyle{alpha}

\end{document}